\documentclass[10pt,letterpaper]{article}
\usepackage[T1]{fontenc}
\usepackage{lmodern}
\usepackage[margin=1in]{geometry}
\usepackage{amsmath,amssymb,booktabs,graphicx}
\usepackage[table]{xcolor}
\usepackage{capt-of}
\usepackage[authoryear,round]{natbib}
\usepackage[hidelinks]{hyperref}
\hypersetup{pdftitle={M-WAM: Motion-Centric World Action Model}}

\title{\bfseries M-WAM: Motion-Centric World Action Model}
\author{
  Guangyang Wu\textsuperscript{1,*}\quad Dingyi Zhao\textsuperscript{1,*}\quad Xiaohong Liu\textsuperscript{1,4,\textdagger}\quad Xiufeng Song\textsuperscript{1}\quad Qi Li\textsuperscript{2}\\[3pt]
  Qiuhong Shen\textsuperscript{2}\quad Guangzhao Li\textsuperscript{1}\quad Yueming Lyu\textsuperscript{3}\quad Guangtao Zhai\textsuperscript{1}\quad Xinchao Wang\textsuperscript{2}\\[6pt]
  {\small\textsuperscript{1}Shanghai Jiao Tong University\enspace\textsuperscript{2}National University of Singapore\enspace\textsuperscript{3}A*STAR\enspace\textsuperscript{4}Shanghai Innovation Institute}
}
\date{}

\begin{document}
\raggedbottom
\maketitle
\begingroup
\renewcommand{\thefootnote}{\fnsymbol{footnote}}
\footnotetext[1]{Equal contribution.}
\footnotetext[2]{Corresponding author.}
\endgroup
\begin{abstract}
World-action models learn robot control alongside future prediction, but their predictive objectives commonly emphasize visual appearance. We investigate dense 3D scene motion as a supervision target for action learning. We introduce M-WAM, which predicts pixel-aligned 3D point trajectories organized into image-like motion maps. A pretrained image-generation backbone models these trajectories, while motion and action objectives jointly optimize the scene features used by the policy. In training, we reconstruct dense trajectories on demand from the per-frame simulator geometry. This avoids precomputing every frame pair and reduces export and storage costs from quadratic to linear in episode length. At deployment, the policy generates actions directly without predicting future motion. In RoboTwin ablations, 3D motion improves success by 10.9 percentage points over last-frame RGB prediction and 6.9 points over 2D motion. M-WAM's mean success rate on LIBERO-Plus is 1.7 points higher than ImageWAM's. Across three real-robot tasks, it achieves twice the mean success rate of LingBot-VA. These results support explicit 3D motion prediction as an effective training objective for manipulation policies. All code and data will be publicly released.
Project page: \url{https://whuzhaodingyi.github.io/M-WAM/}.
\end{abstract}

\begin{center}
    \includegraphics[width=\textwidth]{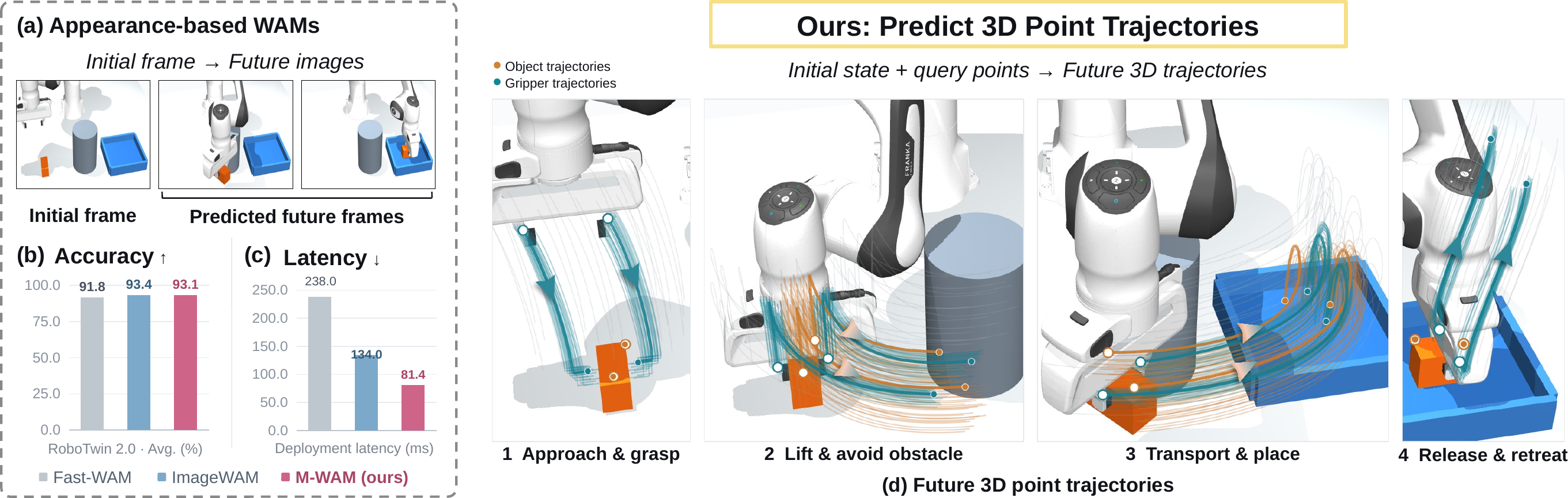}
    \captionof{figure}{\textbf{From appearance prediction to 3D motion supervision.} (a) Future-image prediction in appearance-based WAMs. (b,c) Mean RoboTwin~2.0 success and model forward-pass latency for Fast-WAM~\citep{fastwam}, ImageWAM~\citep{imagewam}, and M-WAM (Tables~\ref{tab:exp-main} and~\ref{tab:exp-efficiency}). (d) Illustration of object (orange) and gripper (teal) trajectories through approach, obstacle avoidance, placement, and retreat.}
    \label{fig:overview}
\end{center}
\section{Introduction}

Vision--language--action (VLA) models transfer visual and language priors to robot control~\citep{zitkovich2023rt,kim2024openvla,black2024pi_0,bjorck2025gr00t,shukor2025smolvla,team2025gemini}. Video-based policies incorporate visual plans~\citep{du2023learninguniversalpoliciestextguided,zhou2024robodreamer,du2024video,bharadhwaj2024gen2act} or predictive features~\citep{hu2024video,feng2025vidar,pai2025mimicvideo}, while world--action models (WAMs) jointly learn future observation prediction and action generation~\citep{li2025unified,zhu2025unifiedworldmodelscoupling,dreamzero,cheang2024gr2generativevideolanguageactionmodel,cen2025WorldVLA}. Predictive co-training can benefit control without generating future observations at deployment~\citep{fastwam}, making the prediction target a central design choice.

Recent WAMs predict future RGB observations, visual latents~\citep{lingbot-va2026,bi2025motusunifiedlatentaction}, or edited images~\citep{imagewam}. These targets combine object motion with texture, lighting, and background appearance. Visual reconstruction leaves 3D point displacement implicit. Yet a small displacement relative to the gripper can determine manipulation success while leaving most pixels unchanged. This motivates our question: \textit{can dense 3D scene motion provide a more effective prediction target for action learning?}

Point trajectories explicitly represent scene displacement~\citep{generalflow,huang2026pointworld,lee2026mu_0}, including motion along the viewing direction and spatial relations between objects and the gripper. Prior models incorporate geometry through depth~\citep{zhen2025tesseract,x-wam}, pointmaps~\citep{xu2026kinema4d}, future-depth readouts~\citep{li2026wam4d}, or trajectory-based feature alignment~\citep{fourDwam}. We instead make \textbf{dense 3D trajectories the generative future-prediction target in joint motion--action learning}.

We introduce M-WAM, which jointly learns dense 3D motion prediction and robot action generation (Figure~\ref{fig:overview}(d)). Pixel-aligned, image-like motion maps let a pretrained PixelDiT~\citep{pixeldit} backbone model object and gripper trajectories. Each source pixel stores the future 3D trajectory of its corresponding scene point. Motion and action objectives jointly optimize current-scene features, which the policy combines with language and proprioception to generate actions without predicting future motion at deployment.

We reconstruct trajectory targets on demand from per-frame simulator geometry under a piecewise-rigid scene model. The same records support different source frames and temporal offsets, reducing simulator export and offline storage from quadratic to linear in episode length compared with precomputing all frame pairs.

Experiments on RoboTwin~2.0~\citep{robotwin2}, LIBERO~\citep{liu2023libero}, LIBERO-Plus~\citep{liberoplus}, and three real-robot tasks evaluate both the prediction target and the resulting policy. M-WAM achieves competitive results on standard RoboTwin and LIBERO evaluations and exceeds ImageWAM by 1.7 points on LIBERO-Plus. Figure~\ref{fig:overview}(b,c) summarizes RoboTwin success and model forward-pass latency. We further evaluate motion--action consistency and robustness to camera shifts.

Our contributions are threefold:

\noindent\textbf{$\diamond$\quad Motion-centric world--action modeling.} We introduce M-WAM, which represents future scene change as pixel-aligned, source-frame 3D point trajectories. Motion prediction shapes the scene representation used for action generation without requiring generated motion at inference.

\noindent\textbf{$\diamond$\quad  Scalable 3D motion supervision.} We reconstruct requested trajectories on demand from per-frame geometry under a piecewise-rigid scene model, avoiding pairwise flow rendering and keeping simulator export and offline storage linear in trajectory length.

\noindent\textbf{$\diamond$\quad  Extensive validation.} In RoboTwin ablations, 3D motion improves success by 10.9 percentage points over last-frame RGB and 6.9 over 2D motion. M-WAM also exceeds ImageWAM's mean success on LIBERO-Plus and doubles LingBot-VA's across three real-robot tasks.

\section{Related Work}

\subsection{Vision--Language--Action and World--Action Models}

VLA models adapt vision--language backbones to robot control~\citep{zitkovich2023rt,kim2024openvla,black2024pi_0,intelligence2025pi_,bjorck2025gr00t,wen2025dexvla}; video-based policies incorporate generated plans~\citep{du2023learninguniversalpoliciestextguided,zhou2024robodreamer,liang2024dreamitate,ko2024learning} or predictive features~\citep{feng2025vidar,hu2024video}. WAMs jointly learn future visual dynamics and actions~\citep{li2025unified,zhu2025unifiedworldmodelscoupling,dreamzero,cen2025WorldVLA,cen2025rynnvla,won2025dualstreamdiffusionworldmodelaugmented}. Related approaches transfer video-generation pretraining to control~\citep{wu2023unleashing,cheang2024gr2generativevideolanguageactionmodel,kim2026cosmos,liang2025videogenerators}. Latent prediction also supervises policy representations~\citep{zheng2025flarerobotlearningimplicit,sun2026vla}, while predicted visual goals guide action generation~\citep{zhao2025cotvlavisualchainofthoughtreasoning}. Fast-WAM retains the benefits of predictive co-training without generating future observations at deployment~\citep{fastwam}. ImageWAM replaces video generation with instruction-conditioned image editing and conditions actions on editing representations~\citep{imagewam}. Their prediction targets emphasize visual appearance or latent change; M-WAM instead predicts dense 3D motion to train scene features for action generation.

\subsection{3D Motion and Geometric World Modeling}

Policies incorporate spatial and motion cues through spatial encoding~\citep{qu25spatialvla,yuan2025depthvla}, visual traces~\citep{zheng2025tracevla}, and feature alignment~\citep{li2025spatial}. General Flow uses dense motion as a reusable affordance representation~\citep{generalflow}. TesserAct predicts RGB, depth, and surface-normal sequences~\citep{zhen2025tesseract}, PointWorld models manipulation dynamics through 3D point flow~\citep{huang2026pointworld}, and Kinema4D uses kinematic 4D pointmaps~\citep{xu2026kinema4d}. These methods model geometry or motion without using dense trajectories as the prediction target of a jointly trained world--action policy.

Within WAMs, WAM4D adds geometric supervision through spatial register tokens~\citep{li2026wam4d}, X-WAM adds RGB-D prediction and 3D reconstruction~\citep{x-wam}, and 4D-WAM aligns features with trajectory fields while retaining visual prediction~\citep{fourDwam}. $\mu_0$ predicts sparse semantic 3D interaction traces for a separate action expert~\citep{lee2026mu_0}. M-WAM instead uses dense, pixel-aligned 3D trajectories as the generative prediction target, rather than an auxiliary geometric readout, feature-alignment signal, or sparse semantic trace. Motion and action objectives jointly train the scene representation used by the policy.

\begin{figure}[!t]
    \centering
    \includegraphics[width=0.985\textwidth]{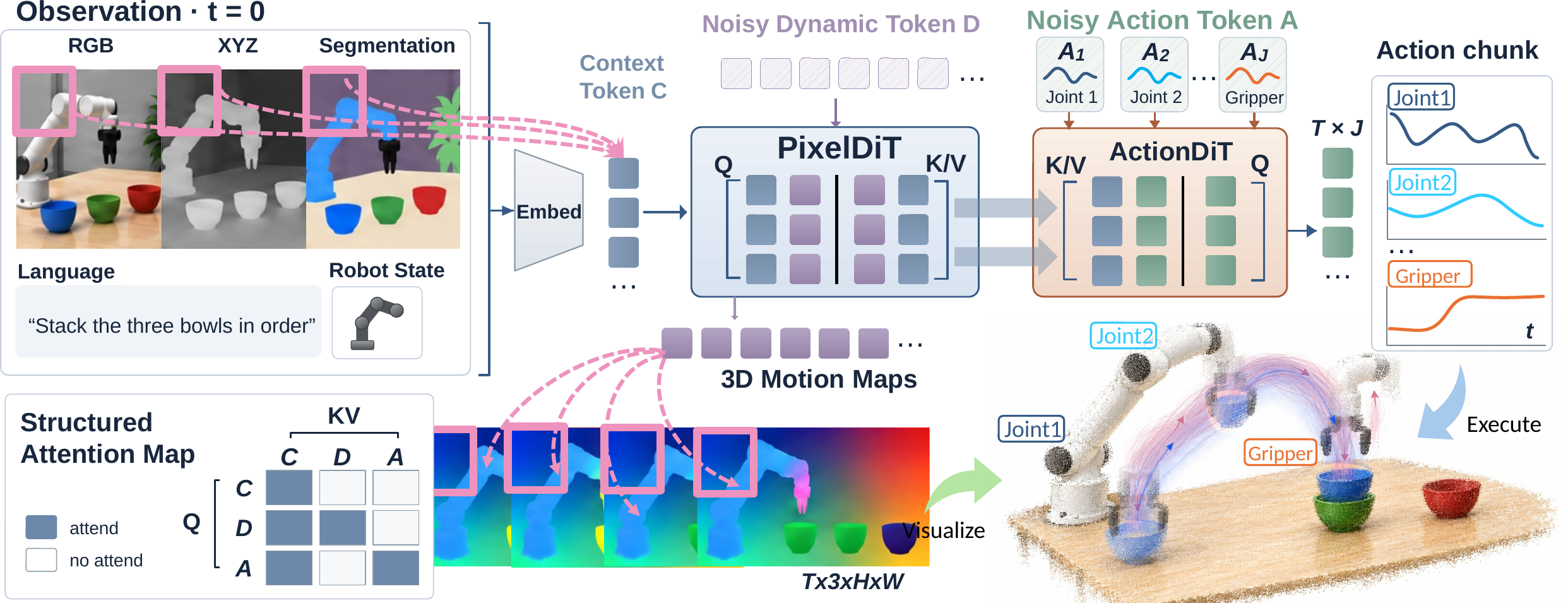}
    \caption{\textbf{M-WAM architecture.} M-WAM takes current RGB, XYZ, and segmentation observations, a language instruction, and the robot state as inputs. PixelDiT predicts pixel-aligned 3D motion maps, with $T$ future 3D displacement vectors stored at each source pixel. ActionDiT uses $J$ tokens, one per control dimension; each token encodes a full $T$-step trajectory, producing a $T\times J$ action chunk. The attention mask allows dynamic (D) and action (A) tokens to read context (C) tokens, while preventing action tokens from reading dynamic tokens.}
    \label{fig:mwam-architecture}
\end{figure}

\section{Method}
\label{sec:method}

\subsection{Overview and Problem Formulation}
\label{sec:mwam-overview}
\label{sec:mwam-3-1}

We consider language-conditioned robotic manipulation from current multiview observations. At time $t$, the model receives
\begin{equation}
O_t=
\left\{
I_t^v,\,\mathbf{P}_t^v,\,\mathbf{S}_t^v
\right\}_{v\in\mathcal{V}},
\label{eq:mwam-observations}
\end{equation}
where $\mathcal{V}$ is the set of available views, and $I_t^v$, $\mathbf{P}_t^v$, and $\mathbf{S}_t^v$ denote the RGB observation, per-pixel XYZ coordinates, and segmentation encoding of view $v$, respectively. The XYZ coordinates are expressed in the source reference frame defined in Section~\ref{sec:mwam-motion-representation}. The formulation does not assume a fixed number or semantic type of cameras. Together with the robot proprioceptive state $\mathbf{q}_t$ and task instruction $l$, the policy predicts an action chunk
\begin{equation}
\mathbf{A}_t=
\left(\mathbf{a}_t,\ldots,\mathbf{a}_{t+H-1}\right)
\in\mathbb{R}^{H\times J},
\qquad
\mathbf{A}_t\sim
\pi_\theta(\cdot\mid O_t,l,\mathbf{q}_t),
\label{eq:mwam-policy}
\end{equation}
where $H$ is the action horizon and $J$ is the number of control dimensions.

M-WAM jointly learns future scene motion and robot action from the current scene (Figure~\ref{fig:mwam-architecture}). Let $Z_C=E_C(O_t)$ denote condition tokens extracted from the current observations. The dynamics (D) and action (A) branches predict flow-matching velocity fields for a noisy 3D motion target and a noisy action chunk at noise levels $\sigma_D$ and $\sigma_A$, respectively:
\begin{equation}
\widehat{\mathbf{v}}_{D}=F_D\!\left(\widetilde{\mathbf{M}}_{t,\sigma_D},Z_C,l,\sigma_D\right),
\qquad
\widehat{\mathbf{v}}_{A}=F_A\!\left(\widetilde{\mathbf{A}}_{t,\sigma_A},Z_C,l,\mathbf{q}_t,\sigma_A\right).
\label{eq:mwam-denoisers}
\end{equation}
The action branch does not consume a generated motion map. Instead, motion prediction serves as a world-modeling objective that shapes the condition representation on which action generation relies. At deployment, actions are generated directly from the current observations, language instruction, and proprioceptive state.

\subsection{Source-Frame 3D Motion Maps}
\label{sec:mwam-motion-representation}

We represent future scene change with pixel-aligned 3D point trajectories. For each sampled source frame $t$, we designate one observation view $r\in\mathcal{V}$ as the reference view and fix its camera coordinate system at time $t$ as the source reference frame $\mathcal{F}_t^{\mathrm{ref}}$. All current XYZ maps and future 3D positions in the sample are expressed in this frame. No shared simulator coordinate convention is required across platforms.

For source pixel $p$ in view $v$, let $\mathbf{x}_{t\rightarrow s}^{v}(p)\in\mathbb{R}^{3}$ denote the position, expressed in $\mathcal{F}_t^{\mathrm{ref}}$, of the physical point selected by $(v,p)$ at source time $t$ when that point is transported to time $s$. The input XYZ maps are expressed in the same source reference frame.
For a sampled clip, let $\Delta_t=(\delta_1,\ldots,\delta_K)$ denote its ordered sequence of positive future offsets. These offsets may be fixed within a training configuration or changed across temporal settings without changing the stored simulation data. The motion target is
\begin{equation}
\mathbf{M}_t^v(p,k)
=
\mathbf{x}_{t\rightarrow t+\delta_k}^{v}(p)
-
\mathbf{x}_{t\rightarrow t}^{v}(p),
\qquad
k=1,\ldots,K.
\label{eq:mwam-canonical-motion}
\end{equation}
The notation anchors point identity at the source frame: $p$ does not denote a pixel in a future observation. Under the piecewise-rigid scene model used to construct simulation supervision, the same physical point can be transported geometrically to future times even when it becomes occluded or leaves the camera view. The supervision therefore tracks represented source points through future occlusion.

For each view, the $K$ displacement vectors are stored at their source-pixel locations, yielding
\begin{equation}
\mathbf{M}_t^v
\in
\mathbb{R}^{h_v\times w_v\times 3K}.
\label{eq:mwam-motion-map-shape}
\end{equation}
Here, $h_v$ and $w_v$ are the image height and width of view $v$. Each output pixel represents the future 3D trajectory of its corresponding source point, while patch-level dynamic tokens encode full-horizon motion over their associated spatial regions. We use $\mathbf{M}_t$ to denote the collection of per-view motion maps in one training sample; operations on $\mathbf{M}_t$ are applied to all available views. XYZ inputs and 3D motion targets are standardized independently using training-set statistics for each dataset. This routine normalization accommodates differences in coordinate ranges and motion scales across platforms.

\subsection{Motion-Supervised World--Action Model}
\label{sec:mwam-architecture}
\label{sec:mwam-3-3}

M-WAM uses an image-generation backbone for 3D motion prediction and a diffusion-transformer action branch for control. The image-like motion maps allow the dynamics branch to build on a pretrained PixelDiT~\citep{pixeldit}. We adapt the input embeddings for RGB, normalized XYZ, segmentation, and noisy motion, together with the output head for the $3K$-channel motion target. The remaining backbone components follow PixelDiT.

\subsubsection{Asymmetric Condition--Dynamics Attention}
\label{sec:mwam-information-flow}

Within the dynamics branch, condition tokens $Z_C$ and noisy dynamic tokens $Z_D$ form a joint sequence
\begin{equation}
Z_{CD}=[Z_C;Z_D],
\end{equation}
but use asymmetric masked self-attention. The subscripts C, D, and A identify condition, dynamic, and action tokens, respectively; $Q$, $K$, and $V$ denote their query, key, and value projections. Suppressing the layer index, the attention sublayer computes
\begin{equation}
Z_C^{+}=\operatorname{Attn}(Q_C,K_C,V_C),
\qquad
Z_D^{+}=\operatorname{Attn}\!\left(Q_D,[K_C;K_D],[V_C;V_D]\right).
\label{eq:mwam-cw-attention}
\end{equation}
These equations describe the attention outputs; other transformer operations are omitted. Condition tokens therefore remain independent of noisy dynamic tokens in the forward pass, whereas dynamic tokens use the current scene to predict future motion. Because $Z_D^{+}$ depends on keys and values computed from $Z_C$, the motion-prediction loss backpropagates through the condition representation. The asymmetric mask thus lets motion supervision shape scene features without allowing sampled or noisy future motion to contaminate them.

\subsubsection{Trajectory-Wise Action Representation}
\label{sec:mwam-action-tokenization-section}

The action branch represents an action chunk by control dimension rather than by individual time step. For the $j$-th control dimension, one token encodes its complete $H$-step trajectory:
\begin{equation}
\mathbf{z}_{A,j}
=
E_A\!\left(
\widetilde{\mathbf{A}}_{t,\sigma_A}[:,j]
\right)
+
\mathbf{e}_j,
\qquad
j=1,\ldots,J,
\label{eq:mwam-action-tokenization}
\end{equation}
where $E_A:\mathbb{R}^{H}\rightarrow\mathbb{R}^{d_{\mathrm{emb}}}$ is a learnable projection, $d_{\mathrm{emb}}$ is the action-token embedding dimension, and $\mathbf{e}_j$ identifies the control dimension. Let $Z_A=[\mathbf{z}_{A,1};\ldots;\mathbf{z}_{A,J}]$. Self-attention among these tokens models coordination across control dimensions while each token retains the full temporal evolution of one control variable. At corresponding layers, action queries read action and condition features:
\begin{equation}
Z_A^{+}
=
\operatorname{Attn}
\left(
Q_A,
[K_C;K_A],
[V_C;V_A]
\right).
\label{eq:mwam-action-attention}
\end{equation}
The action branch does not read dynamic tokens. Its scene context is provided by condition features jointly optimized by the motion and action objectives. A final projection maps the action-token features to an $H\times J$ flow-matching velocity prediction. Language, proprioception, and noise-level conditioning are applied throughout the action branch.

\subsubsection{Training Strategies}
\label{sec:mwam-training-strategies}
\label{sec:mwam-training-objective}

Following Fast-WAM~\citep{fastwam}, we train both branches with flow matching~\citep{flowmatching}, independently perturbing the motion and action targets and minimizing the sum of their mean squared velocity errors.

\paragraph{High-noise-biased sampling.}
Motion maps and action chunks contain strong low-frequency trajectory structure. During inference, the earliest denoising steps start close to pure noise and establish the coarse global layout of the predicted trajectories. We therefore reweight the noise-level distribution for both branches toward this regime. Specifically, the training noise-level distribution $p_\sigma^{\mathrm{train}}$ satisfies
$\Pr_{\sigma\sim p_\sigma^{\mathrm{train}}}[\sigma\in[0.93,1]]=0.20$.
The remaining probability mass is renormalized over lower noise levels. This increases supervision where the model must recover global motion and action structure from nearly pure noise, while retaining coverage of lower-noise refinement steps.

\paragraph{Robust geometric and segmentation conditioning.}
The XYZ conditions are constructed from depth, but we perturb the input depth during training before back-projection and transformation to the source reference frame. Only the conditioning geometry is perturbed; motion supervision is constructed from the unperturbed recorded geometry. This exposes the policy to imperfect geometric observations while preserving accurate trajectory targets. Segmentation conditions are generated automatically from RGB observations using Grounding DINO and SAM~2 rather than simulator-provided ground-truth masks.

\subsection{Scalable Motion Supervision from Per-Frame Geometry}
\label{sec:mwam-data-pipeline}
\label{sec:mwam-3-2}

The motion target in Eq.~\eqref{eq:mwam-canonical-motion} can be generated without rendering pair-specific flow. Simulator-specific exporters convert each episode into a common per-frame representation containing metric depth, camera intrinsics and poses, per-pixel rigid-entity associations, and the poses of rigid objects and robot links. Static scene regions are represented as entities with fixed poses. The entity associations are used only to construct supervision and are not provided to the policy.

Consider a source pixel $p$ observed by camera $v$ at time $t$, and let $b=b_t^v(p)$ denote its associated rigid entity. For readability, we omit $(v,p)$ from intermediate point variables. Let $T^{\mathrm{cam}}_v(t)$ and $T^{\mathrm{cam}}_r(t)$ map coordinates from the source camera and the designated reference camera, respectively, to simulator coordinates, and let $T^{\mathrm{ent}}_b(s)$ map entity-local coordinates to simulator coordinates at time $s$. All transforms are in $\mathrm{SE}(3)$; homogeneous-coordinate notation is omitted. We first back-project the source pixel to simulator coordinates,
\begin{equation}
\mathbf{y}_t
=
T^{\mathrm{cam}}_v(t)\,
\operatorname{BackProj}
\left(p,d_t^v(p);\mathbf{K}_t^v\right),
\label{eq:mwam-backprojection}
\end{equation}
where $d_t^v(p)$ and $\mathbf{K}_t^v$ are the stored depth and camera intrinsics. The same physical point is then transported with its rigid entity and expressed in the reference-camera coordinate system fixed at source time $t$:
\begin{equation}
\mathbf{x}_{t\rightarrow s}
=
T^{\mathrm{cam}}_r(t)^{-1}
T^{\mathrm{ent}}_b(s)T^{\mathrm{ent}}_b(t)^{-1}
\mathbf{y}_t.
\label{eq:mwam-rigid-transport}
\end{equation}
Here, $T^{\mathrm{ent}}_b(t)^{-1}$ maps the source point into the entity frame, $T^{\mathrm{ent}}_b(s)$ moves it to time $s$, and $T^{\mathrm{cam}}_r(t)^{-1}$ expresses the result in $\mathcal{F}_t^{\mathrm{ref}}$. Thus, $\mathbf{x}_{t\rightarrow s}$ is precisely $\mathbf{x}_{t\rightarrow s}^{v}(p)$ defined in Section~\ref{sec:mwam-motion-representation}. Setting $s=t+\delta_k$ gives the motion target $\mathbf{M}_t^v(p,k)=\mathbf{x}_{t\rightarrow t+\delta_k}-\mathbf{x}_{t\rightarrow t}$. Because the point is propagated through entity motion rather than recovered from a target image, its trajectory remains defined through future occlusion and does not depend on target-frame visibility.

For an $N$-frame trajectory, pre-rendering pairwise flow requires $O(NK)$ targets for a fixed set of $K$ offsets and can grow to $O(N^2)$ when arbitrary source--target pairs must be supported. Our pipeline instead stores one geometry record per frame and constructs only the motion targets requested by the dataloader. Simulator-side export and offline geometry storage therefore remain $O(N)$, independent of the temporal offsets used for training or ablation.

\definecolor{mwamrow}{RGB}{221,237,236}
\makeatletter
\newcommand{\mwamfullrowshade}{%
  \noalign{%
    \begingroup
    \color{mwamrow}%
    \hrule height\dimexpr\ht\@arstrutbox+\dp\@arstrutbox\relax
    \vskip-\dimexpr\ht\@arstrutbox+\dp\@arstrutbox\relax
    \endgroup
  }%
}
\makeatother
\section{Experiments}
\label{sec:experiments}
We evaluate dense 3D motion supervision through manipulation benchmarks, predictive-target comparisons, visual perturbations, and motion diagnostics. Architectural and training ablations examine which design choices contribute to action performance and inference efficiency.

\subsection{Experimental Setup}
\label{sec:exp-setup}
\paragraph{Sim \& Real Benchmark.}
We evaluate on RoboTwin~2.0~\citep{robotwin2}, LIBERO~\citep{liu2023libero}, LIBERO-Plus~\citep{liberoplus}, and an ARX5 dual-arm robot. RoboTwin training uses 20,000 demonstrations across 50 tasks, with 200 episodes per task under each of the Clean and Randomized conditions. All RoboTwin closed-loop success evaluations, including ablations, follow the Fast-WAM protocol with 10,000 test episodes: 100 per task per condition, evaluated with RoboTwin's task-specific success checks. LIBERO covers the Spatial, Object, Goal, and Long suites. LIBERO-Plus evaluates the same trained policies under seven perturbation categories, using only the original LIBERO demonstrations for training. The ARX5 tasks are block insertion (T1), cube stacking (T2), and drawer loading (T3), each with 200 training demonstrations and 20 evaluation trials under randomized initial object placements. We construct real-robot 3D motion supervision from depth and point trajectories estimated using VGGT-$\Omega$ and SpatialTrackerV2.

\paragraph{Baseline \& Metrics.}
We compare with Fast-WAM~\citep{fastwam}, Motus~\citep{bi2025motusunifiedlatentaction}, LingBot-VA~\citep{lingbot-va2026}, and ImageWAM~\citep{imagewam}. Table~\ref{tab:exp-main} reports unweighted averages of the displayed component scores. ImageWAM reports 2,500 Clean and 25,000 Randomized RoboTwin training demonstrations, whereas M-WAM uses the balanced 20,000-episode set described above. The LIBERO-Plus comparison uses ImageWAM's FLUX.2 4B variant. We report success rate (SR) in percent and absolute differences in percentage points. For motion evaluation, Avg.~$\delta$~\citep{tapvid} averages tracking accuracy over error thresholds and future frames, using ground-truth-visible points. Endpoint error (EPE) measures the mean Euclidean distance between predicted and demonstration endpoints. Motion--Action Discrepancy (MAD) measures the same distance using endpoints induced by executing the corresponding action chunk as the reference. EPE and MAD include all queried points, irrespective of endpoint visibility. Both distances are measured in image-plane pixels; lower is better. For each fixed model, motion metrics are reported as mean $\pm$ standard deviation over repeated sampling with different random noise initializations.

\subsection{Manipulation Performance}
\label{sec:exp-main}
\paragraph{Simulation performance.}
Table~\ref{tab:exp-main} compares performance across the three simulation benchmarks. On RoboTwin~2.0 Clean, M-WAM achieves the highest success rate, exceeding ImageWAM by 1.70 percentage points. Averaged over Clean and Randomized conditions, M-WAM reaches 93.10\%, 1.27 points above Fast-WAM and 0.28 points below ImageWAM. On LIBERO, its 98.3\% average is within 0.2 points of the highest-scoring baseline, LingBot-VA.

The larger visual and layout variations in LIBERO-Plus produce clearer differences between methods. M-WAM achieves an average of 84.6\%, improving over ImageWAM by 1.7 points and Fast-WAM by 33.1 points. It exceeds ImageWAM in five of the seven categories, with the largest gains on Robot (+9.6 points) and Background (+4.4 points). ImageWAM retains higher scores on Light and Noise. Thus, methods with similar success on standard LIBERO exhibit substantially different performance under the broader LIBERO-Plus evaluation.

\paragraph{Real-robot performance.}
M-WAM achieves the highest success rate on all three ARX5 tasks (Figure~\ref{fig:exp-real} and Table~\ref{tab:exp-real}). It succeeds in 8, 11, and 9 of the 20 trials for block insertion, cube stacking, and drawer loading, respectively. On each task, this is four more successful trials, or 20 percentage points, than the strongest baseline. Its mean success rate is 46.7\%, twice the 23.3\% achieved by LingBot-VA, the strongest baseline on average.

\paragraph{Inference efficiency.}
We measure model forward-pass costs on RoboTwin inputs using an NVIDIA RTX 4090. Latency covers the model computation needed to produce an action chunk from prepared inputs; observation preprocessing is outside the timed region. M-WAM has the lowest compute cost, latency, and peak memory among the methods in Table~\ref{tab:exp-efficiency}. Its forward pass requires 673.84 GFLOPs and 81.39 ms, with 3.54 GB peak memory. Fast-WAM and ImageWAM take 2.92 and 1.65 times as long, respectively.

\begin{table}[!t]
\centering
\caption{\textbf{Success on simulation benchmarks (\%, $\uparrow$).} Best and second-best results are bold and underlined; shaded rows denote our method. ``-'' denotes unavailable results.}
\label{tab:exp-main}
\scriptsize
\renewcommand{\arraystretch}{1.08}
\setlength{\tabcolsep}{1.5pt}
\resizebox{\linewidth}{!}{%
\begin{tabular}{@{}lccc@{\hspace{6pt}}ccccc@{\hspace{6pt}}cccccccc@{}}
\toprule
& \multicolumn{3}{c}{RoboTwin 2.0} & \multicolumn{5}{c}{LIBERO} & \multicolumn{8}{c}{LIBERO-Plus} \\
\cmidrule(lr){2-4}\cmidrule(lr){5-9}\cmidrule(lr){10-17}
Method & Clean & Rand. & Avg. & Spa. & Obj. & Goal & Long & Avg. & Cam. & Robot & Lang. & Light & Bkg. & Noise & Layout & Avg. \\
\midrule
Fast-WAM & 91.88 & \underline{91.78} & 91.83 & \underline{98.2} & \textbf{100.0} & 97.0 & 95.2 & 97.6 & 16.4 & 44.5 & 68.9 & 78.2 & 53.7 & 37.7 & 60.7 & 51.4 \\
Motus & 88.66 & 87.02 & 87.84 & 96.8 & \underline{99.8} & 96.6 & 97.6 & 97.7 & - & - & - & - & - & - & - & - \\
LingBot-VA & 92.90 & 91.50 & 92.20 & \textbf{98.5} & 99.6 & 97.2 & \textbf{98.5} & \textbf{98.5} & 13.8 & 35.7 & 24.1 & 28.6 & 20.7 & 21.4 & 13.8 & 22.6 \\
ImageWAM & \underline{93.20} & \textbf{93.56} & \textbf{93.38} & 97.2 & 99.2 & \textbf{98.8} & \underline{98.4} & \underline{98.4} & \underline{80.8} & \underline{50.3} & \underline{91.4} & \textbf{98.1} & \underline{85.5} & \textbf{93.8} & \underline{80.5} & \underline{82.9} \\
\mwamfullrowshade
M-WAM (ours) & \textbf{94.90} & 91.30 & \underline{93.10} & 97.8 & 99.0 & \underline{97.9} & \textbf{98.5} & 98.3 & \textbf{81.0} & \textbf{59.9} & \textbf{93.8} & \underline{94.3} & \textbf{89.9} & \underline{90.2} & \textbf{83.0} & \textbf{84.6} \\
\bottomrule
\end{tabular}%
}
\end{table}

\begin{table}[!t]
\begin{minipage}[t]{0.405\linewidth}
\vspace{0pt}\centering
\caption{\raggedright\textbf{Denoising steps and SR (\%).}}
\label{tab:exp-denoise}
\scriptsize
\setlength{\tabcolsep}{1pt}
\begin{tabular*}{\linewidth}{@{\extracolsep{\fill}}lcccccc@{}}
\toprule
Denoising steps & 1 & 2 & 3 & 5 & 10 & 20 \\
\midrule
SR (\%) & 88.21 & 89.57 & 91.35 & \textbf{93.10} & \underline{92.03} & 91.06 \\
\bottomrule
\end{tabular*}

\caption{\raggedright\textbf{Predictive targets.}}
\label{tab:exp-representation}
\scriptsize
\setlength{\tabcolsep}{1.6pt}
\begin{tabular*}{\linewidth}{@{\extracolsep{\fill}}lcc@{}}
\toprule
Target & Success Rate \% $\uparrow$ & MAD $\downarrow$ \\
\midrule
Last-frame RGB & 82.17 & 4.23 \textpm{} 0.89 \\
2D motion & \underline{86.22} & \underline{2.42} \textpm{} 0.41 \\
\mwamfullrowshade
3D motion (ours) & \textbf{93.10} & \textbf{1.15} \textpm{} 0.07 \\
\bottomrule
\end{tabular*}
\end{minipage}\hfill
\begin{minipage}[t]{0.575\linewidth}
\vspace{0pt}\centering
\caption{\raggedright\textbf{Predicted and executed motion.} Avg.~$\delta$ (\%); EPE and MAD (px), mean $\pm$ SD.}
\label{tab:exp-motion}
\scriptsize
\renewcommand{\arraystretch}{1.1}
\setlength{\tabcolsep}{0.6pt}
\begin{tabular*}{\linewidth}{@{\extracolsep{\fill}}lcccc@{}}
\toprule
& \multicolumn{2}{c}{Prediction} & \multicolumn{2}{c}{Alignment} \\
\cmidrule(lr){2-3}\cmidrule(l){4-5}
Method & Avg.~$\delta$ $\uparrow$ & EPE $\downarrow$ & Avg.~$\delta$ $\uparrow$ & MAD $\downarrow$ \\
\midrule
Fast-WAM & 91.26 \textpm{} 0.27 & \underline{1.48} \textpm{} 0.04 & \textbf{91.66} \textpm{} 0.35 & \underline{1.26} \textpm{} 0.06 \\
Motus & \underline{91.55} \textpm{} 0.21 & 1.67 \textpm{} 0.11 & 90.65 \textpm{} 0.26 & 1.32 \textpm{} 0.06 \\
LingBot-VA & 89.26 \textpm{} 0.33 & 3.85 \textpm{} 0.09 & 88.19 \textpm{} 0.64 & 5.32 \textpm{} 0.55 \\
ImageWAM & N/A & 3.51 \textpm{} 0.06 & N/A & 1.76 \textpm{} 0.07 \\
\mwamfullrowshade
M-WAM  & \textbf{93.06} \textpm{} 0.20 & \textbf{1.01} \textpm{} 0.02 & \underline{91.64} \textpm{} 0.30 & \textbf{1.15} \textpm{} 0.07 \\
\bottomrule
\end{tabular*}
\end{minipage}
\end{table}

\subsection{Which Future Representation Supports Action?}
\label{sec:exp-representation}
We isolate the predictive target by keeping the input modalities, model, and training setup fixed and comparing last-frame RGB, 2D motion, and 3D motion (Table~\ref{tab:exp-representation}). Dense 3D motion achieves the highest success rate among these targets: 93.10\%, improving over last-frame RGB by 10.93 percentage points and 2D motion by 6.88 points. Prediction--execution agreement improves alongside task success: MAD decreases from 4.23 pixels for RGB to 2.42 for 2D motion and 1.15 for 3D motion. Relative to 2D motion, the 3D target reduces MAD by 52.5\%, supporting its use for learning both successful actions and motion predictions that agree with their execution.

\subsection{Robustness to Camera and Lighting Changes}
\label{sec:exp-robustness}
\begin{figure}[!t]
\centering
\begin{minipage}[t]{0.51\linewidth}
\vspace{0pt}\centering
\scriptsize
\setlength{\tabcolsep}{1.8pt}
\begin{tabular*}{0.97\linewidth}{@{\extracolsep{\fill}}lccc@{}}
\toprule
Method & Ref. & Camera change & Low light \\
\midrule
Fast-WAM & 91.83 & \underline{52.64} (\textminus{}39.19) & 80.46 (\textminus{}11.37) \\
Motus & 87.84 & 41.45 (\textminus{}46.39) & 76.10 (\textminus{}11.74) \\
LingBot-VA & 92.20 & 35.67 (\textminus{}56.53) & 85.11 (\textminus{}7.09) \\
ImageWAM & \textbf{93.38} & 47.58 (\textminus{}45.80) & \underline{86.36} (\textminus{}7.02) \\
\mwamfullrowshade
M-WAM (ours) & \underline{93.10} & \textbf{86.46} (\textminus{}6.64) & \textbf{87.01} (\textminus{}6.09) \\
\bottomrule
\end{tabular*}
\end{minipage}\hfill
\begin{minipage}[t]{0.47\linewidth}
\vspace{3pt}\centering
\includegraphics[width=\linewidth]{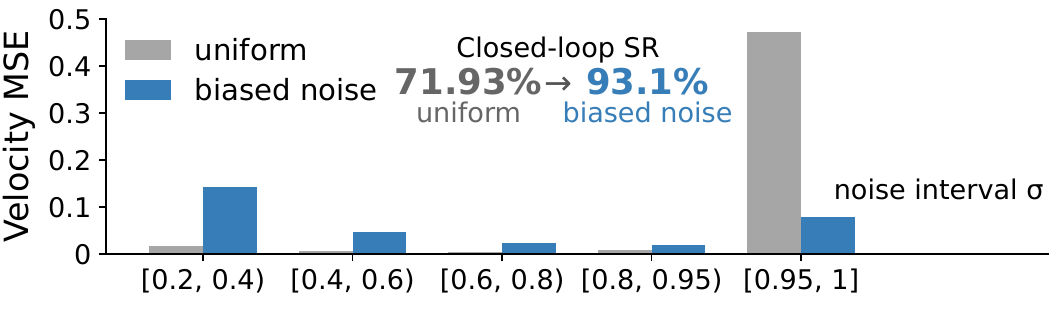}
\par
\end{minipage}
\par\nointerlineskip
\begin{minipage}[t]{0.51\linewidth}
\vspace{0pt}\centering
\captionof{table}{\centering\textbf{RoboTwin robustness (SR, \%).}}
\label{tab:exp-shifts}
\end{minipage}\hfill
\begin{minipage}[t]{0.47\linewidth}
\vspace{0pt}\centering
\caption{\centering\textbf{Noise sampling ablation.}}
\label{fig:exp-noise}
\end{minipage}

\end{figure}

We test robustness by modifying observations in the original RoboTwin evaluation episodes while retaining the same trained policies, initial states, and task dynamics. In Table~\ref{tab:exp-shifts}, Ref. denotes performance on the unmodified test conditions. For the perturbed evaluations, we raise the camera rig by 5 cm or scale image brightness by 0.3. After the camera shift, we recompute the calibration used for depth back-projection and express XYZ in the moved head-camera coordinate frame. Parenthesized values report changes in success rate, in percentage points, relative to each method's unperturbed reference.

M-WAM achieves the highest success rate and the smallest decrease under both perturbations. Under the camera shift, its success decreases by 6.64 percentage points to 86.46\%, whereas the baselines lose 39.19--56.53 points. M-WAM exceeds the strongest baseline under this shift, Fast-WAM, by 33.82 points. Under low light, M-WAM achieves 87.01\%, exceeding ImageWAM, the strongest baseline, by 0.65 points. Its decrease of 6.09 points is also smaller than the baseline drops of 7.02--11.74 points.

\begin{figure}[!t]
\centering
\includegraphics[width=\linewidth]{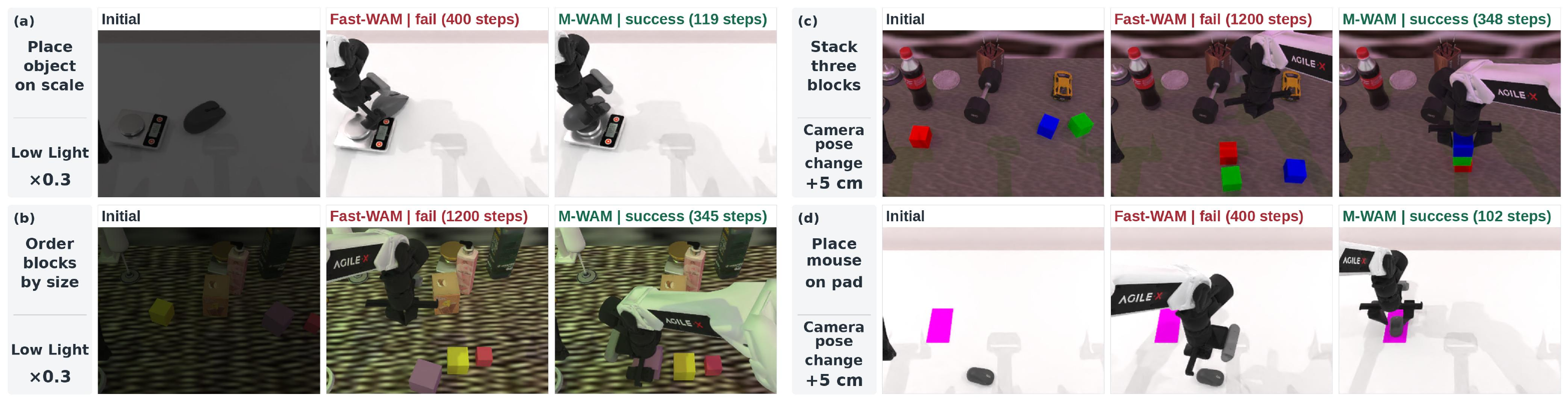}

\caption{\textbf{Selected rollouts under visual perturbations.} (a,b) Brightness $\times0.3$. (c,d) Camera raised by 5 cm. Each triplet shows the initial observation, a failed Fast-WAM rollout, and a successful M-WAM rollout from the same initial state. Headers report rollout step counts.}
\label{fig:exp-qualitative}
\end{figure}

Figure~\ref{fig:exp-qualitative} illustrates the effect of these perturbations on individual rollouts. Across the four selected cases, M-WAM completes the task while Fast-WAM fails from the same initial state, under either reduced brightness or a shifted camera.

\begin{figure}[!t]
\centering
\begin{minipage}[t]{0.44\linewidth}
\centering
\captionof{table}{\raggedright\textbf{Real-robot success rate.} ARX5 dual-arm robot; 20 trials per task.}
\label{tab:exp-real}
\scriptsize
\setlength{\tabcolsep}{1.2pt}
\begin{tabular*}{\linewidth}{@{\extracolsep{\fill}}lcccc@{}}
\toprule
Method & T1 & T2 & T3 & Avg. \\
\midrule
Motus~\citep{bi2025motusunifiedlatentaction} & 0\% & 15\% & 5\% & 6.7\% \\
Fast-WAM~\citep{fastwam} & \underline{20\%} & 25\% & 20\% & 21.7\% \\
ImageWAM~\citep{imagewam} & 10\% & 20\% & 20\% & 16.7\% \\
LingBot-VA~\citep{lingbot-va2026} & 10\% & \underline{35\%} & \underline{25\%} & \underline{23.3\%} \\
\mwamfullrowshade
M-WAM (ours) & \textbf{40\%} & \textbf{55\%} & \textbf{45\%} & \textbf{46.7\%} \\
\bottomrule
\end{tabular*}
\end{minipage}\hfill
\begin{minipage}[t]{0.54\linewidth}
\vspace{0pt}\centering
\includegraphics[width=\linewidth]{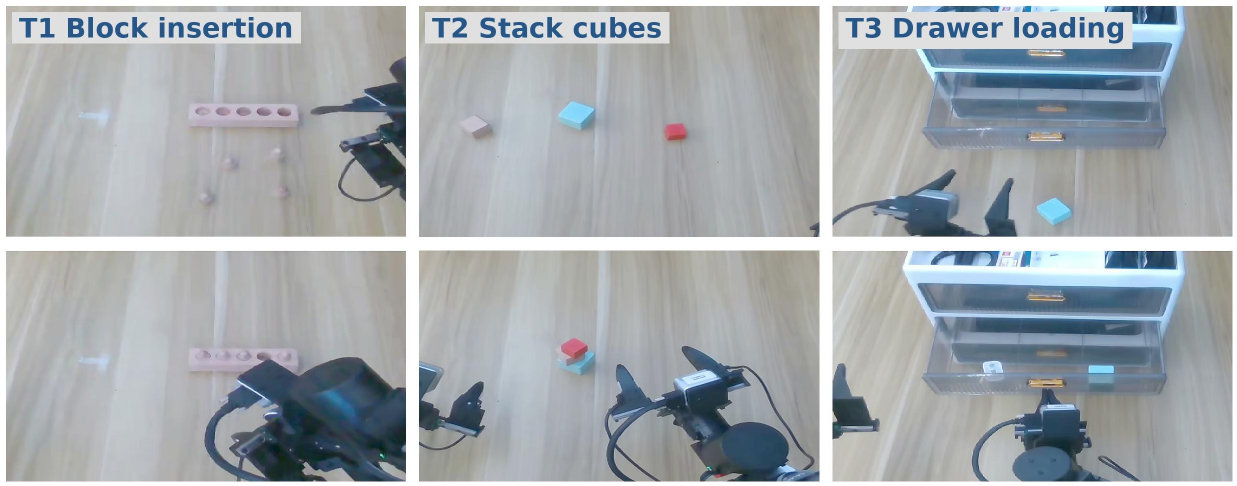}
\caption{The three ARX5 manipulation tasks.}\label{fig:exp-real}
\end{minipage}
\end{figure}

\subsection{Do Predicted and Executed Motions Agree?}
\label{sec:exp-alignment}
For prediction accuracy, we compare generated motion with held-out demonstration trajectories. For prediction--execution alignment, we start from the same simulator state and execute a single generated action chunk over the prediction horizon, without replanning. Execution endpoints are obtained from simulator geometry. Queries cover all source-image points. We project simulator trajectories and predicted 3D trajectories into the image plane, and use CoTracker3~\citep{cotracker3} to extract point correspondences from RGB predictions. For ImageWAM, CoTracker3 establishes correspondences between the current and predicted endpoint frames; we therefore report EPE and MAD, with full-horizon Avg.~$\delta$ marked N/A. For policies deployed without future prediction, we enable the predictive branch for this diagnostic while retaining the same action output.

M-WAM achieves the highest prediction Avg.~$\delta$ and the lowest prediction EPE in Table~\ref{tab:exp-motion}. Relative to Fast-WAM, its full-horizon prediction accuracy improves by 1.80 percentage points, and its endpoint EPE decreases from 1.48 to 1.01 pixels, a 31.8\% reduction.

For alignment with executed actions, M-WAM achieves the lowest MAD: 1.15 pixels, compared with 1.26 for Fast-WAM, an 8.7\% reduction. Their full-horizon Avg.~$\delta$ scores are nearly identical (91.64\% and 91.66\%). M-WAM therefore improves motion forecasting and endpoint agreement with executed actions, while full-horizon alignment remains comparable to Fast-WAM.

\begin{table}[!tb]
\begin{minipage}[t]{0.345\linewidth}
\vspace{0pt}\centering
\caption{\raggedright\textbf{Inference cost.}}
\label{tab:exp-efficiency}
\scriptsize
\setlength{\tabcolsep}{1.2pt}

\begin{tabular*}{\linewidth}{@{\extracolsep{\fill}}lccc@{}}
\toprule
Method & GFLOPs & ms & GB \\
\midrule
Fast-WAM & \underline{2849.30} & 238.00 & 13.66 \\
ImageWAM & 4838.24 & \underline{133.96} & \underline{10.13} \\
\mwamfullrowshade
M-WAM & \textbf{673.84} & \textbf{81.39} & \textbf{3.54} \\
\bottomrule
\end{tabular*}
  
\end{minipage}\hfill
\begin{minipage}[t]{0.32\linewidth}
\vspace{0pt}\centering
\caption{\raggedright\textbf{Attention  ablation.}}
\label{tab:exp-attention}
\scriptsize
\setlength{\tabcolsep}{1.2pt}
\vspace{0.5\baselineskip}
\begin{tabular*}{\linewidth}{@{\extracolsep{\fill}}llcc@{}}
\toprule
Attn. & Future & SR & ms \\
\midrule
$C\leftrightarrow D$ & Joint & \underline{89.14} & \underline{128.44} \\
\mwamfullrowshade
$C\rightarrow D$ & Omit & \textbf{93.10} & \textbf{81.39}\\
\bottomrule
\end{tabular*}
\end{minipage}\hfill
\begin{minipage}[t]{0.305\linewidth}
\vspace{0pt}\centering
\caption{\raggedright\textbf{Action tokenization.} }
\label{tab:exp-tokens}
\scriptsize
\setlength{\tabcolsep}{1.2pt}
\vspace{0.5\baselineskip}
\begin{tabular*}{\linewidth}{@{\extracolsep{\fill}}lcc@{}}
\toprule
Tokens & SR & MAD $\downarrow$ \\
\midrule
Time-wise       & \underline{91.52} & \underline{2.72} \textpm{} 0.41 \\
\mwamfullrowshade
Trajectory-wise & \textbf{93.10} & \textbf{1.15} \textpm{} 0.07 \\
\bottomrule
\end{tabular*}
\end{minipage}

\end{table}

\subsection{Architectural and Training Ablations}
\label{sec:exp-ablations}
We evaluate the architectural and training choices on the full RoboTwin~2.0 test set, using the same 10,000-episode protocol as the main evaluation.

\paragraph{Attention ablation.}
The attention structure determines whether action inference depends on future denoising. With bidirectional attention ($C\leftrightarrow D$), context features depend on noisy dynamic tokens, requiring joint future and action generation. One-way conditioning ($C\rightarrow D$) prevents dynamic tokens from feeding back into the context used by the action branch, allowing future denoising to be omitted. Table~\ref{tab:exp-attention} compares these two structures with their corresponding inference procedures. The asymmetric model achieves 93.10\% success, 3.96 percentage points above the bidirectional model, while reducing latency from 128.44 to 81.39 ms (36.6\%).

\paragraph{Action tokenization.}
Trajectory-wise tokenization assigns one token to each of the $J$ control dimensions, with each token encoding a complete $H$-step trajectory. Time-wise tokenization instead assigns one token to each of the $H$ time steps, encoding a $J$-dimensional action. In Table~\ref{tab:exp-tokens}, trajectory-wise tokenization improves success by 1.58 percentage points and reduces MAD from 2.72 to 1.15 pixels, a 57.7\% reduction.

\paragraph{Denoising steps.}
Five denoising steps achieve the highest success (Table~\ref{tab:exp-denoise}). Success rises from 88.21\% with one step to 93.10\% with five, a gain of 4.89 percentage points. Increasing to 10 or 20 steps does not improve success.

\paragraph{High-noise-biased sampling.}
Figure~\ref{fig:exp-noise} plots motion flow-matching velocity MSE by noise level. Uniform sampling produces a sharp error increase at $\sigma\in[0.95,1]$. Motion is largely smooth across space and time and dominated by low-frequency structure; finer high-frequency details are harder to reconstruct. At low and moderate noise, much of the dominant structure remains recoverable, making these cases relatively easy. Uniform sampling therefore allocates most updates to easier cases, which can leave near-pure-noise denoising undertrained. The biased strategy in Section~\ref{sec:mwam-training-strategies} reduces high-noise error, although errors at lower noise levels increase. Closed-loop success improves from 71.93\% to 93.10\%, a gain of 21.17 percentage points.

\section{Conclusion}
We introduced M-WAM, which jointly learns robot actions and dense 3D point trajectories represented as pixel-aligned motion maps. Motion supervision shapes the scene features used by the policy, enabling direct action generation without predicting future motion at deployment. Trajectories are reconstructed on demand from per-frame geometry, keeping simulator export and offline storage linear in episode length. Target ablations and motion--action diagnostics support this choice of supervision, while simulation and real-robot evaluations demonstrate its utility for manipulation. These findings highlight the prediction target as an important design choice for world--action models and establish dense 3D motion as an effective supervision signal for action learning.

\section*{Acknowledgments}
The work was supported by the National Natural Science Foundation of China under Grant 62572317.

\bibliographystyle{plainnat}
\bibliography{references}
\end{document}